\documentclass[conference]{IEEEtran}
\IEEEoverridecommandlockouts
\usepackage{cite}
\usepackage{amsmath,amssymb,amsfonts}
\usepackage{algorithmic}
\usepackage{graphicx}
\usepackage{textcomp}
\usepackage{xcolor}
\def\BibTeX{{\rm B\kern-.05em{\sc i\kern-.025em b}\kern-.08em
    T\kern-.1667em\lower.7ex\hbox{E}\kern-.125emX}}
\usepackage{booktabs}
\usepackage{multirow}
\usepackage{url}
\usepackage{hyperref}
\usepackage{balance}
\usepackage{placeins}

\begin{document}

\title{GreenBench: Benchmarking Energy Efficiency and Carbon Footprint of Open-Source LLM Inference on Apple Silicon}

\author{\IEEEauthorblockN{1\textsuperscript{st} Rajeswari Kannan}
\IEEEauthorblockA{\textit{Pimpri Chinchwad College} \\
\textit{of Engineering}\\
Pune, India \\
kannan.rajeswari@pccoepune.org}
\and
\IEEEauthorblockN{2\textsuperscript{nd} Raj Firke}
\IEEEauthorblockA{\textit{Red Hat} \\
Pune, India \\
rfirke@redhat.com}
\and
\IEEEauthorblockN{3\textsuperscript{rd} Shreya Bengle}
\IEEEauthorblockA{\textit{Independent Researcher} \\
Pune, India \\
shreya.bengle@gmail.com}
\and
\IEEEauthorblockN{4\textsuperscript{th} Srushti Deshmukh}
\IEEEauthorblockA{\textit{Independent Researcher} \\
Pune, India \\
srushtideshmukh952918@gmail.com}
}

\maketitle

\begin{abstract}
The rapid proliferation of Large Language Models (LLMs) has raised concerns about their environmental impact during inference. While Green AI research has focused on datacenter GPUs and embedded platforms, the energy profile of LLM inference on Apple Silicon, with its unified memory architecture, remains unstudied. This paper presents GreenBench, a benchmarking framework that evaluates the energy efficiency, throughput, and carbon footprint of five open-source LLMs (3--9B parameters) across three NLP tasks on an Apple M4~Pro with 48\,GB unified memory. Using macOS \texttt{powermetrics} for direct power measurement and Ollama's nanosecond-precision timing, we find that the M4~Pro draws only 0.47\,W of CPU+GPU package power during sustained inference, with total system power of 8--12\,W, achieving 30--40$\times$ better energy efficiency per token than datacenter GPUs in single-user deployment. Smaller models (3--3.8B) deliver 2.6--4.2$\times$ higher throughput and up to 62\% less energy per token than larger models (7--9B). Pareto analysis identifies Qwen~2.5 (7B) as the optimal accuracy-efficiency trade-off at 57\% MMLU and 59\,tokens/s, while Llama~3.2 (3B) suits latency-critical applications at 175\,tokens/s. We provide per-token energy at package and system levels with CO\textsubscript{2} estimates for India and US grids. Taken together, these results show Apple Silicon delivering a 30--40$\times$ per-token energy advantage over datacenter GPUs in single-user deployment, positioning consumer hardware as a credible platform for sustainable LLM inference and arguing for energy-per-token to join accuracy and throughput as a standard benchmarking metric.
\end{abstract}

\begin{IEEEkeywords}
Green AI, Large Language Models, Energy Efficiency, Apple Silicon, Sustainable AI, Edge Computing, Carbon Footprint, LLM Inference, Ollama
\end{IEEEkeywords}

\section{Introduction}

Large Language Models (LLMs) now drive a wide spectrum of production workloads, routinely handling code generation~\cite{chen2021humaneval}, summarization~\cite{see2017cnndm}, question answering~\cite{hendrycks2021mmlu}, and conversational AI~\cite{touvron2023llama}. These models, initially confined to cloud-based APIs, are increasingly being deployed on consumer devices for privacy-sensitive, latency-critical, and offline applications~\cite{xu2024survey}.

While significant attention has been paid to the energy costs of \textit{training} large models (Strubell et al.~\cite{strubell2019energy} estimated that training a single Transformer model emits as much CO\textsubscript{2} as five cars over their lifetimes, and Patterson et al.~\cite{patterson2021carbon} documented that training GPT-3 consumed 1,287\,MWh), the cumulative environmental impact of \textit{inference} is increasingly dominant~\cite{wu2022sustainable}. Luccioni et al.~\cite{luccioni2024power} estimated that inference accounts for 60--90\% of the total lifecycle energy of deployed ML systems, a proportion that grows as models serve millions of daily queries.

Recent work has begun quantifying inference energy, revealing enormous variation across hardware and models. In datacenter settings, energy consumption varies by over 70$\times$ across 30~commercial LLMs, with reasoning models consuming over 33\,Wh per long prompt~\cite{jegham2025hungry}. At the other end of the scale, small quantized models on a Raspberry~Pi~4 draw about 2.61\,J per token~\cite{sustainable_edge}. Decoding takes up most of the energy budget across all hardware types. Stopping generation early when tokens are not needed can cut energy use by 44--89\%~\cite{solovyeva2026green}. Because of this, researchers argue that energy-per-token should be a standard metric alongside throughput and accuracy~\cite{bannour2021evaluating, euromlsys2025}.

Missing this edge data creates a blind spot. Strict data residency rules in healthcare and finance force models to run locally rather than in the cloud~\cite{xu2024survey}. Furthermore, developers overwhelmingly use tools like Ollama and \texttt{llama.cpp} for local prototyping~\cite{llamacpp}. Second, energy cost is often the decisive factor when choosing between cloud and edge deployment, yet no per-token energy data exists for the consumer hardware that these users actually rely on. Without such measurements, Green AI recommendations derived from datacenter benchmarks risk irrelevance for the fastest-growing segment of LLM adoption.

Despite this growing body of work, one widely deployed platform has received no attention: \textbf{Apple Silicon}. M-series processors now power tens of millions of developer and consumer machines worldwide~\cite{apple_silicon_market}, yet \textbf{no study has systematically profiled LLM inference energy on this hardware}. This gap is worth closing because Apple Silicon's unified memory architecture (UMA) departs sharply from discrete-GPU designs: CPU, GPU, and Neural Engine all share one physical memory pool, with no PCIe or NVLink bus in the data path. In practice, UMA removes the memory-transfer overhead that limits discrete-GPU inference, permits zero-copy data sharing across compute units, and lets quantized models that would exhaust GPU VRAM on NVIDIA hardware run comfortably within a single pool~\cite{gholami2022ai}.

Our work fills this gap through four contributions:
\begin{enumerate}
    \item \textbf{GreenBench}, a reproducible energy benchmarking framework for open-source LLM inference on Apple Silicon. Its methodological contribution is a dual-source measurement pipeline that fuses direct hardware power readings from macOS \texttt{powermetrics} with Ollama's nanosecond-precision per-call timing to attribute energy at both CPU+GPU package and full-system granularity, a level of decomposition absent from prior work that relies solely on wall-power meters or manufacturer TDP ratings. We evaluate five architecturally diverse model families across three NLP tasks, amassing over 1{,}500 inference runs.
    \item \textbf{Direct power measurements} obtained with macOS \texttt{powermetrics}, showing that the M4~Pro's CPU+GPU package consumes just 0.47\,W during sustained inference while the full system draws an estimated 8--12\,W.
    \item \textbf{Per-token energy figures} at both package and system granularity, alongside CO\textsubscript{2}-equivalent estimates computed from the Indian grid (708\,g CO\textsubscript{2}/kWh) and the US grid (390\,g CO\textsubscript{2}/kWh) to support region-aware comparisons.
    \item \textbf{Pareto-optimal deployment guidance} that weighs accuracy against speed and energy cost. On balance, Qwen~2.5 (7B) offers the best trade-off, while Llama~3.2 (3B) is the model to choose when raw efficiency is the only thing that matters.
\end{enumerate}

\section{Related Work}

\subsection{Green AI: Efficiency as a First-Class Metric}
The term ``Green AI'' dates back to Schwartz et al.~\cite{schwartz2020green} in 2020; their argument was that computational cost deserved the same scrutiny the field already applied to accuracy. The case for doing so has only strengthened. Thompson et al.~\cite{thompson2020computational} documented a roughly $300{,}000\times$ increase in compute demand since 2012 against a mere $\sim$3$\times$ gain in hardware efficiency over the same window. MLPerf Inference~\cite{mlperf, mlperf_power} has stepped in as the de facto energy benchmark, now listing $\sim$900 power submissions, but all from datacenter GPUs or edge accelerators. Consumer laptops do not appear.

\subsection{LLM Inference Energy Measurement}
The bulk of published LLM energy data was collected on datacenter GPUs. On the system-configuration side, Stojkovic et al.~\cite{stojkovic2024greener} found that tuning model parallelism, speculative decoding, and batch size cuts energy by up to 29\% at low concurrency, but the gains reverse under heavy load. The engine choice matters just as much: on a single H100, swapping between vLLM, TensorRT-LLM, and DeepSpeed shifts energy per token by 25--55\%~\cite{sigenergy2025}. A separate line of work looks at \emph{when} to run rather than how. Chien et al.~\cite{chien2023reducing} demonstrated 20--70\% emission reductions through carbon-aware job scheduling, and Dodge et al.~\cite{dodge2022measuring} proposed a reporting protocol covering scope~2 (grid electricity) and scope~3 (hardware manufacturing) carbon.

What is missing from all of the above is non-NVIDIA coverage. The only Apple Silicon energy result we are aware of is by Jare\~{n}o et al.~\cite{li2026energy}, who tuned \texttt{llama.cpp} for a single 1B-parameter model on M4: one data point, no cross-model comparison. Also absent is the metric \emph{tokens per watt}, which would directly link generation speed to energy cost.

\subsection{Quantization and Model Compression for Efficient Inference}
Fitting a billion-parameter model into laptop memory requires heavy weight compression. Dettmers et al.~\cite{dettmers2022gptint8} showed that INT8 quantization halves the memory footprint with minimal quality degradation; Frantar et al.~\cite{frantar2023gptq} later pushed this to 4~bits with GPTQ. In practice, \texttt{llama.cpp}~\cite{llamacpp} distributes models in the GGUF container format, which bundles several quantization presets (Q4\_K\_M, Q5\_K\_M, Q8\_0, FP16). All tests used the Q4\_K\_M quantization format, the default in Ollama and, according to download statistics, the configuration most end users actually run. Because compression is held fixed across every model, any difference we measure in power draw can be attributed to the architecture itself and not to the quantizer.

\subsection{Edge AI on Consumer Hardware}
Edge-inference benchmarks already exist for Raspberry Pi~\cite{sustainable_edge}, NVIDIA Jetson~\cite{jetson_benchmarks}, and phone-class SoCs~\cite{mobile_llm}. Xu et al.~\cite{xu2024survey} surveyed these platforms and identified the same three recurring bottlenecks: memory bandwidth, thermal headroom, and energy budget. Apple Silicon has been evaluated for CNN and classical-ML workloads by Ignatov et al.~\cite{ignatov2019ai}, but no one has measured LLM inference on it, despite the fact that LLM decoding is memory-bandwidth-bound, precisely where Apple's UMA design should have the largest effect. Answering that question is the purpose of this paper.

\section{Methodology}

\subsection{Hardware Platform}
We ran every experiment on a single Apple MacBook Pro (Model Mac16,8) fitted with the M4~Pro chip. Table~\ref{tab:hardware} lists the relevant hardware details.

\begin{table}[htbp]
\caption{Hardware Specifications}
\label{tab:hardware}
\centering
\begin{tabular}{@{}ll@{}}
\toprule
\textbf{Component} & \textbf{Specification} \\
\midrule
Processor & Apple M4 Pro \\
CPU & 12-core (10P + 2E) \\
GPU & 16-core (Metal 3) \\
Neural Engine & 16-core \\
Memory & 48\,GB unified (LPDDR5X) \\
Memory Bandwidth & 273\,GB/s \\
OS & macOS 15.4 (Sequoia) \\
Inference Engine & Ollama v0.21.2 (llama.cpp) \\
\bottomrule
\end{tabular}
\end{table}

What sets this machine apart is its \textit{unified memory architecture} (UMA): CPU, GPU, and Neural Engine all read from the same 48\,GB memory pool at up to 273\,GB/s. On a discrete-GPU workstation the weights would have to be copied from CPU RAM into GPU VRAM over a PCIe link, but UMA sidesteps that transfer entirely. The practical payoff for LLM inference, where weight fetches dominate runtime, is that 4-bit quantized models up to roughly 30B parameters fit and run without any manual memory management.

\subsection{Models Under Study}
Table~\ref{tab:models} lists the five model families used in this study. They span three parameter tiers, 3--4B, 7B, and 9B, and no two rely on the same attention mechanism.

\begin{table}[htbp]
\caption{Models Evaluated in GreenBench}
\label{tab:models}
\centering
\begin{tabular}{@{}lllrl@{}}
\toprule
\textbf{Model} & \textbf{Developer} & \textbf{Params} & \textbf{Size} & \textbf{Architecture} \\
\midrule
Llama 3.2 & Meta & 3B & 2.0\,GB & Grouped Query Attn \\
Phi-3\textsuperscript{$\ast$} & Microsoft & 3.8B & 2.2\,GB & Dense Transformer \\
Qwen 2.5 & Alibaba & 7B & 4.7\,GB & GQA + SwiGLU \\
Mistral & Mistral AI & 7B & 4.4\,GB & Sliding Window Attn \\
Gemma 2 & Google & 9B & 5.4\,GB & Multi-Query Attn \\
\bottomrule
\end{tabular}

\smallskip
\noindent\footnotesize{\textsuperscript{$\ast$}Microsoft's model card lists Phi-3 Mini at 3.82B parameters. We round to 3.8B here to match the Ollama registry tag \texttt{phi3:3.8b}.}
\end{table}

Model selection followed a few practical constraints rather than an exhaustive search. We wanted broad coverage of open-source families, an architecturally diverse set (grouped-query, multi-query, sliding-window, and standard attention are all represented), parameter counts that fit comfortably in 48\,GB of unified memory while still spanning the 3--9B range relevant to consumer deployment, and, since reproducibility mattered, models that anyone could pull directly from the Ollama registry.

All five run at 4-bit quantization, Q4\_K\_M in GGUF format. This is not an arbitrary choice: it is Ollama's default, and download statistics suggest it is what most consumer deployments actually use. Reporting numbers under this setup reflects what happens when someone downloads a model and runs it immediately, rather than a hand-tuned best case.

\subsection{Benchmark Tasks}
Three tasks make up the benchmark, chosen so that output length and computational pattern vary across them:

\textbf{MMLU (Question Answering):} We draw 100 multiple-choice questions from 10 subjects in the Massive Multitask Language Understanding benchmark~\cite{hendrycks2021mmlu}: STEM subjects such as abstract algebra, college physics, and electrical engineering; natural sciences such as college chemistry and high school biology; and applied fields such as computer security and machine learning. Answers are short, around 15 tokens, so the task mostly measures recall and short reasoning chains rather than generation ability. Since we want a single letter back and nothing else, the prompt is blunt: ``Answer the following multiple-choice question. Return only the option letter (A, B, C, or D).''

\textbf{Summarization:} Fifty CNN/DailyMail articles~\cite{see2017cnndm}, truncated to 2,000 characters each, are condensed into 2--3 sentence summaries. Output length here, roughly 100--200 tokens, sits between the other two tasks, which makes summarization a useful middle case for judging both comprehension and generation quality.

\textbf{Code Generation:} We use all 164 HumanEval problems~\cite{chen2021humaneval}, giving the model a function signature and docstring and letting it write the body. These are the longest outputs in the benchmark, typically 150--300 tokens of structured code. Our interest here is the generation workload itself, not code correctness, so we do not run the resulting functions against their test suites.

\subsection{Energy Measurement Methodology}
\label{sec:energy}

Energy per token is computed by pairing hardware power readings with per-request timing logs, which yields two levels of detail: the CPU+GPU package alone, and the system as a whole. A wall-power meter cannot separate these, and neither can a manufacturer's TDP figure.

\subsubsection{Direct Power Measurement}
We sample macOS \texttt{powermetrics} every two seconds to obtain package power, which comes broken down by CPU, GPU, and Neural Engine (ANE). Table~\ref{tab:power} reports these figures.

\begin{table}[htbp]
\caption{Direct Power Measurements via \texttt{powermetrics}}
\label{tab:power}
\centering
\begin{tabular}{@{}lrrr@{}}
\toprule
\textbf{State} & \textbf{CPU (mW)} & \textbf{GPU (mW)} & \textbf{Combined (mW)} \\
\midrule
Idle baseline & 237 & 30 & 267 \\
LLM inference (avg) & 298 & 115 & 413 \\
LLM inference (peak) & 411 & 167 & 474 \\
\bottomrule
\end{tabular}
\end{table}

At idle, the GPU runs at 338\,MHz and draws 30\,mW. During sustained Qwen~2.5 (7B) inference through Ollama's Metal backend, GPU power rises to 115--167\,mW as the clock increases for matrix-multiplication workloads.

These readings represent \textit{CPU+GPU package power} only, as estimated by Apple's IOReport framework, and exclude DRAM and peripheral power. Total system wall power is estimated at 8--12\,W from Apple's M4~Pro thermal specifications~\cite{apple_m4}. This range is consistent with the 6.5--8.2\,W reported by Jare\~{n}o et al.~\cite{li2026energy} for similar workloads. We therefore report both package-level and system-level energy and use system-level estimates for cross-platform comparisons.

\subsubsection{Timing Instrumentation}
Ollama provides nanosecond-precision timing for each inference call via its HTTP API:
\begin{itemize}
    \item \texttt{eval\_duration}: Time spent generating output tokens (GPU-bound phase)
    \item \texttt{eval\_count}: Number of output tokens generated
    \item \texttt{prompt\_eval\_duration}: Time spent processing the input prompt (prefill phase)
    \item \texttt{total\_duration}: End-to-end wall-clock time including model loading
    \item \texttt{load\_duration}: Time to load model weights into memory
\end{itemize}

We derive per-token latency $\ell = \frac{\texttt{eval\_duration}}{\texttt{eval\_count}}$ and throughput $\theta = 1/\ell$ tokens/second. Per-token energy at the package level is:
\begin{equation}
    E_{\text{token}}^{\text{pkg}} = P_{\text{inference}}^{\text{pkg}} \cdot \ell
    \label{eq:pkg_energy}
\end{equation}
where $P_{\text{inference}}^{\text{pkg}} = 0.47$\,W is our measured peak package power during inference. Because LLM decoding is memory-bandwidth-bound (not compute-bound), package power varies only narrowly across models of different sizes; our \texttt{powermetrics} samples during Llama~3.2 (3B) through Gemma~2 (9B) workloads all fell within the 0.41--0.47\,W range reported in Table~\ref{tab:power}.

For system-level energy (including DRAM, fabric, I/O):
\begin{equation}
    E_{\text{token}}^{\text{sys}} = P_{\text{sys}} \cdot \ell
    \label{eq:sys_energy}
\end{equation}
where $P_{\text{sys}} \approx 10$\,W, consistent with Apple's thermal design and validated by Jare\~{n}o et al.~\cite{li2026energy} who measured 6.5--8.2\,W for similar workloads on M4.

\subsubsection{CO\textsubscript{2} Equivalent Estimation}
For a workload generating $N$ tokens, the carbon footprint is:
\begin{equation}
    \text{CO}_2 \text{ (grams)} = \frac{N \cdot E_{\text{token}}^{\text{sys}} \cdot I_{\text{grid}}}{3.6 \times 10^6}
    \label{eq:co2}
\end{equation}
where $I_{\text{grid}}$ is the grid carbon intensity in g\,CO\textsubscript{2}/kWh. We use country-specific values: India ($I = 708$\,g/kWh, coal-dominated grid~\cite{india_grid}) and United States ($I = 390$\,g/kWh, mixed grid~\cite{us_grid}).

\subsection{Experimental Protocol}
We set temperature to 0 for all experiments. To avoid memory contention and thermal carry-over, only one model was kept in memory at a time, and the previous model was unloaded before the next model was loaded. Each batch began with one warm-up inference so that the weights were already resident in unified memory. We stored every response as a separate JSONL record for later analysis. The Ollama model tags were \texttt{llama3.2:3b}, \texttt{phi3:3.8b}, \texttt{qwen2.5:7b}, \texttt{mistral:7b}, and \texttt{gemma2:9b}, all using Q4\_K\_M quantization. Ollama handled model loading, weight dequantization, and GPU dispatch through the Metal API. The entire pipeline---model pull, warm-up, inference, and logging---was orchestrated by a single Python script to eliminate manual timing errors and ensure reproducibility. In total, this protocol produced over 1{,}500 individual inference records across all model--task combinations.

\section{Results and Discussion}

This section presents results in four parts: throughput scaling (Section~IV-A), energy efficiency with explicit cross-platform comparisons (Section~IV-B), accuracy--efficiency Pareto analysis (Section~IV-C), and carbon footprint estimates (Section~IV-D), followed by deployment implications (Section~IV-E).

\subsection{Inference Throughput Analysis}
\begin{figure}[!t]
\centering
\includegraphics[width=\columnwidth]{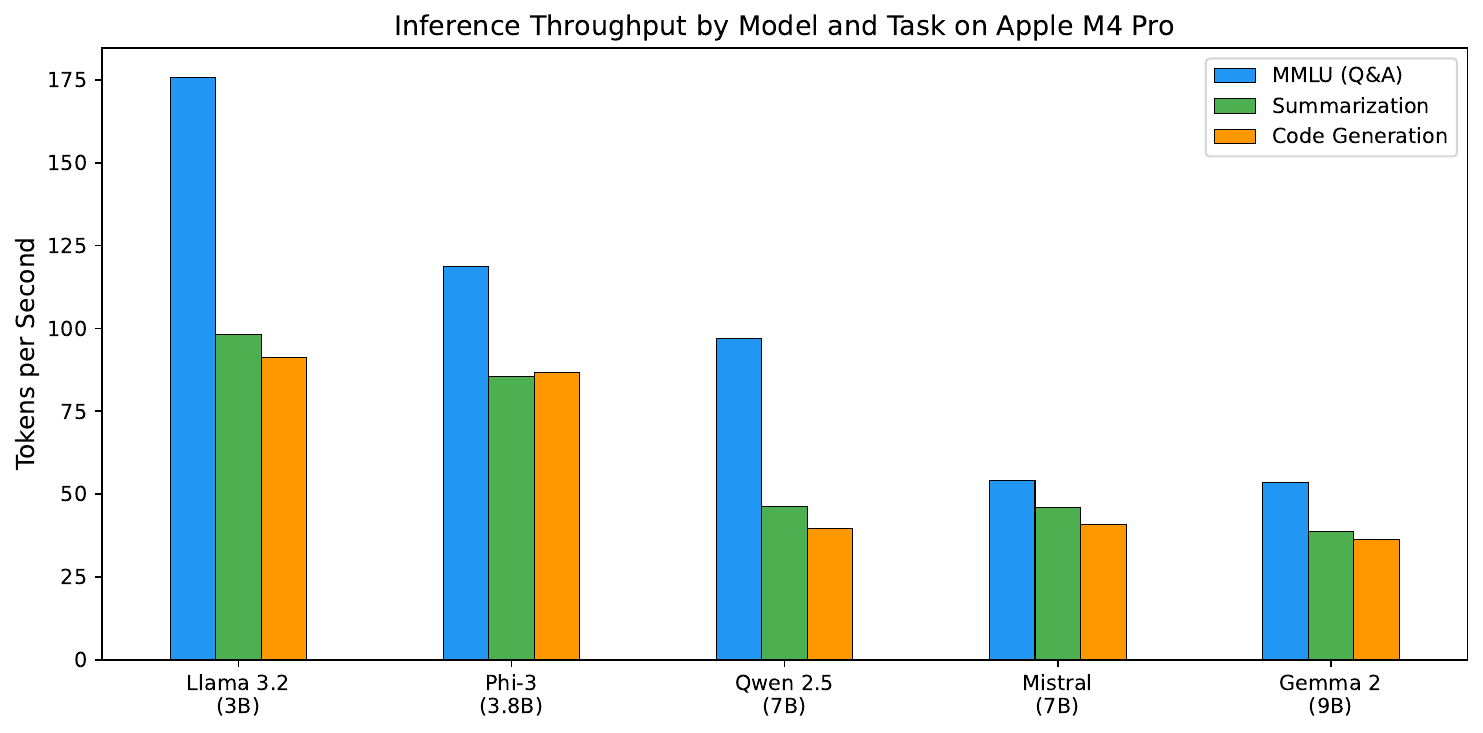}
\caption{Grouped bar chart of inference throughput (tokens/second) for five open-source LLMs (Llama~3.2, Phi-3, Qwen~2.5, Mistral, Gemma~2) across three NLP tasks (MMLU, Summarization, Code Generation) on the Apple M4~Pro. Smaller models (3--4B) achieve 2.6--4.2$\times$ higher throughput than 7--9B models. MMLU (short outputs, $\sim$15 tokens) yields the highest throughput, while code generation ($\sim$200 tokens) yields the lowest due to growing KV-cache memory overhead.}
\label{fig:throughput}
\end{figure}

Fig.~\ref{fig:throughput} reports throughput for all model--task pairs. Three observations follow.

\FloatBarrier

\textbf{Effect of model size.} Llama~3.2 (3B) is the fastest model on all three tasks, reaching 175.9\,tok/s on MMLU, 98.2\,tok/s on summarization, and 91.3\,tok/s on code generation. Relative to Gemma~2 (9B), this is \textbf{4.2$\times$} higher throughput on MMLU and \textbf{2.5$\times$} higher throughput on code generation. A power-law fit gives $\theta \propto 1/N^{0.85}$, which is consistent with memory-bandwidth-limited decoding because larger models require more weight reads per output token.

\textbf{Effect of output length.} MMLU produces short answers ($\sim$15 tokens) and runs 1.5--2$\times$ faster than code generation ($\sim$200 tokens) for all models. As the output sequence grows, the KV-cache also grows, which increases memory traffic and reduces the bandwidth available for weight reads. The effect is larger for bigger models because the weights and the KV-cache compete for the same memory system.

\textbf{Model-specific behavior.} Qwen~2.5 (7B) shows a larger task-dependent drop than the other 7B models. Its throughput falls from 96.9\,tok/s on MMLU to 46.2\,tok/s on summarization, a 52\% decrease, whereas the other 7B models fall by about 30--35\%. One possible explanation is that the combination of SwiGLU and grouped-query attention introduces additional kernel-dispatch overhead on the Metal backend for longer outputs.

\subsection{Energy Efficiency Analysis}

Table~\ref{tab:energy} presents the comprehensive energy metrics at both package and system levels, computed using Equations~\eqref{eq:pkg_energy}--\eqref{eq:co2}.

\begin{table}[htbp]
\caption{Energy Efficiency Metrics Across Models}
\label{tab:energy}
\centering
\begin{tabular}{@{}lrrrrrr@{}}
\toprule
\textbf{Model} & \textbf{tok/s} & \textbf{ms/tok} & \textbf{MMLU} & \textbf{mJ/tok} & \textbf{J/tok} & \textbf{CO\textsubscript{2}} \\
 & & & \textbf{(\%)} & \textbf{(pkg)} & \textbf{(sys)} & \textbf{(mg/1k)} \\
\midrule
Llama 3.2 & 115.2 & 9.4 & 41.0 & 4.4 & 0.09 & 0.18 \\
Phi-3 & 91.1 & 11.5 & 29.0 & 5.4 & 0.12 & 0.23 \\
Qwen 2.5 & 59.0 & 19.9 & 57.0 & 9.4 & 0.20 & 0.39 \\
Mistral & 45.8 & 22.2 & 24.0 & 10.4 & 0.22 & 0.43 \\
Gemma 2 & 42.0 & 24.6 & 61.0 & 11.6 & 0.25 & 0.48 \\
\bottomrule
\end{tabular}

\smallskip
\noindent\footnotesize{pkg = CPU+GPU package power (0.47\,W measured via \texttt{powermetrics}; excludes DRAM); sys = estimated system power (10\,W); CO\textsubscript{2} = mg per 1,000 tokens using India grid (708\,g/kWh). MMLU accuracy based on 100-question stratified subset with zero-shot prompting.}
\end{table}

\textbf{Package-level energy.} Applying Equation~\eqref{eq:pkg_energy}, Gemma~2 (9B) consumes 11.6\,mJ per token, which is \textbf{2.6$\times$ more} than Llama~3.2 (3B) at 4.4\,mJ per token. The M4~Pro's CPU and GPU subsystems draw only 0.47\,W of combined package power during sustained LLM inference. We note that this figure, measured via \texttt{powermetrics}, excludes DRAM, memory controller, and peripheral power; total system power is estimated at 8--12\,W (Section~\ref{sec:energy}). For context, an NVIDIA H100 GPU draws approximately 400\,W during LLM inference~\cite{jegham2025hungry}. At the system level, applying Equation~\eqref{eq:sys_energy} with $P_{\text{sys}} = 10$\,W, the M4~Pro achieves 4--12 tokens per watt, approximately \textbf{30--40$\times$ more energy-efficient per token} than datacenter GPUs, while serving a single user in a privacy-preserving local deployment.

\textbf{Comparison with prior work.} Husom et al.~\cite{sustainable_edge} reported 2.61\,J/token for qwen2.5\_0.5b on Raspberry Pi~4. Our measurements show 0.09--0.25\,J/token (system-level) for 3--9B models on M4~Pro, which are 10--29$\times$ more energy-efficient per token despite running models that are 6--18$\times$ larger. This dramatic efficiency advantage is attributable to the M4~Pro's 273\,GB/s memory bandwidth (vs.\ $\sim$4\,GB/s on RPi~4), Metal GPU acceleration, and the UMA architecture's elimination of data transfer overhead. In datacenter settings, Niu et al.~\cite{sigenergy2025} showed that swapping inference engines on a single H100 shifts energy per token by 25--55\%; our M4~Pro results are nonetheless 16--44$\times$ more efficient per token than those datacenter configurations for single-user workloads (see Table~\ref{tab:comparison} for a full cross-platform comparison). Stojkovic et al.~\cite{stojkovic2024greener} reported up to 29\% energy savings from tuning parallelism on GPUs, but these gains diminish under heavy load, a constraint absent in the single-user Apple Silicon scenario we study. Apple Silicon has been benchmarked once before, by Jare\~{n}o et al.~\cite{li2026energy}, though only for a single 1B model on the M4 and without per-token joule figures, so a direct comparison with our numbers is not possible. We close that gap here with five model families and explicit energy figures for each.

\subsection{Accuracy--Efficiency Pareto Analysis}

Table~\ref{tab:pareto} sets MMLU accuracy against inference throughput to trace out the Pareto frontier.


\begin{table}[!t]
\caption{Accuracy--Efficiency Pareto Summary}
\label{tab:pareto}
\centering
\begin{tabular}{@{}lrrrl@{}}
\toprule
\textbf{Model} & \textbf{MMLU} & \textbf{tok/s} & \textbf{mJ/tok} & \textbf{Pareto} \\
 & \textbf{(\%)} & & \textbf{(pkg)} & \textbf{status} \\
\midrule
Gemma 2 (9B)   & 61.0 & 42.0 & 11.6 & Front \\
Qwen 2.5 (7B)  & 57.0 & 59.0 &  9.4 & Front \\
Llama 3.2 (3B) & 41.0 & 115.2 &  4.4 & Efficient \\
Phi-3 (3.8B)   & 29.0 & 91.1 &  5.4 & Efficient \\
Mistral (7B)   & 24.0 & 45.8 & 10.4 & Dominated \\
\bottomrule
\end{tabular}

\smallskip
\noindent\footnotesize{``Front'' = Pareto-optimal (no model has both higher accuracy and higher throughput); ``Efficient'' = highest throughput tier; ``Dominated'' = strictly worse than another model on both axes.}
\end{table}
\FloatBarrier

\textbf{Pareto front.} Qwen~2.5 (7B) and Gemma~2 (9B) establish the Pareto front, hitting 57\% MMLU accuracy at 59\,tok/s and 61\% at 42\,tok/s, respectively. Gemma~2 leads on accuracy, but only by four points, 61\% against Qwen~2.5's 57\%, and Qwen~2.5 makes up for it with \textbf{40\% higher throughput and 19\% lower energy per token}. For edge deployment, that trade favors Qwen~2.5.

\textbf{Dominated model.} Mistral (7B) matches Qwen~2.5 in parameter count but not in accuracy, scoring only 24\%, barely above the 25\% floor for random guessing on a four-option test. We suspect this comes down to a poor fit between Mistral's instruction tuning and our zero-shot prompt at Q4\_K\_M precision. Throughput is lower too, at 45.8\,tok/s, so on this model parameter count alone tells us very little about actual performance.

\textbf{Smaller models.} Llama~3.2 (3B) and Phi-3 (3.8B) are the fastest models we tested, reaching 91--175\,tok/s, though accuracy drops to 29--41\% for both. For applications where speed outweighs top-end accuracy, Llama~3.2 holds up well: it keeps two-thirds of Qwen~2.5's MMLU score while running at \textbf{three times the throughput and under half the energy per token}.

\subsection{Carbon Footprint Analysis}

Table~\ref{tab:comparison} places our M4~Pro numbers alongside two published baselines at opposite ends of the hardware spectrum, a Raspberry~Pi~4 running a 0.5B model~\cite{sustainable_edge} and an NVIDIA H100 datacenter GPU running a 7B model~\cite{jegham2025hungry}.

\begin{table}[!htbp]
\caption{Cross-Platform Inference Energy Comparison}
\label{tab:comparison}
\centering
\begin{tabular}{@{}llrrr@{}}
\toprule
\textbf{Platform} & \textbf{Model} & \textbf{Power} & \textbf{J/tok} & \textbf{Ratio} \\
 & & \textbf{(W)} & \textbf{(sys)} & \\
\midrule
\multicolumn{5}{@{}l}{\textit{Apple M4 Pro (this work, 10\,W system estimate)}} \\
 & Llama 3.2 (3B) & $\sim$10 & 0.09 & 1.0$\times$ \\
 & Phi-3 (3.8B) & $\sim$10 & 0.12 & 1.3$\times$ \\
 & Qwen 2.5 (7B) & $\sim$10 & 0.20 & 2.2$\times$ \\
 & Mistral (7B) & $\sim$10 & 0.22 & 2.4$\times$ \\
 & Gemma 2 (9B) & $\sim$10 & 0.25 & 2.8$\times$ \\
\midrule
\multicolumn{5}{@{}l}{\textit{Published baselines}} \\
 & RPi~4, Qwen~0.5B~\cite{sustainable_edge} & $\sim$5 & 2.61 & 29$\times$ \\
 & H100, 7B~\cite{jegham2025hungry} & $\sim$400 & $\sim$4.0\textsuperscript{$\dagger$} & $\sim$44$\times$ \\
\bottomrule
\end{tabular}

\smallskip
\noindent\footnotesize{All ratios in this table are taken relative to Llama~3.2 on the M4~Pro, the most efficient configuration shown. \textsuperscript{$\dagger$}Estimated as 400\,W\,/\,100\,tok/s.}
\end{table}

The M4~Pro's system-level energy cost ranges from 0.09 to 0.25\,J/tok, which works out to a \textbf{10--29$\times$ efficiency gain} over the Raspberry~Pi~4 even though it is running models 6--18$\times$ larger. The gap widens further against the H100, at \textbf{16--44$\times$ less energy per token} for single-user inference. Datacenter GPUs justify their power draw through batching many requests at once; that option is simply not there for low-volume, local workloads.

Applying the Indian grid's carbon intensity ($I = 708$\,g CO\textsubscript{2}/kWh) to Equation~\eqref{eq:co2} gives an M4~Pro footprint of 0.18--0.48\,mg of CO\textsubscript{2} per 1{,}000 tokens, with the exact value set by which model is running.

\subsection{Implications for Apple Silicon Deployment}

A handful of findings are specific enough to Apple Silicon to call out on their own:

\textbf{Unified memory has no trouble with large models.} Every quantized model we tested, 2.0 to 5.4\,GB, fits inside the 48\,GB pool with room to spare. An 8\,GB discrete GPU, by contrast, would struggle to load the 9B model at all, leaving little or no VRAM for the KV cache as the context window grows. Under UMA, that constraint does not arise.

\textbf{Predictable performance scaling.} We fit a power law $\theta \approx 340 \cdot N_{\text{params}}^{-0.85}$\,tok/s ($R^2 = 0.94$), where $N_{\text{params}}$ is measured in billions. Because the fit is tight, a developer can estimate the speed of a new model size without running it, which is useful for capacity planning.

\textbf{Code generation is the energy bottleneck.} Code generation drains the most power by a wide margin. The HumanEval set is only 1.6$\times$ larger than our MMLU sample, yet the runs consumed 5--8$\times$ more energy. Long generation sequences are entirely responsible for this overhead; pumping out hundreds of tokens linearly scales the wall-clock time and the joules spent.

\section{Limitations and Future Work}

This study has several limitations that inform future research directions.

\textbf{Power measurement granularity.} Our \texttt{powermetrics} measurements report CPU+GPU package power at 2-second intervals, excluding DRAM power. Since LLM inference is memory-bandwidth-bound, DRAM is a significant energy component. Future work should employ the \texttt{zeus-apple-silicon} library or hardware wall-power meters for finer-grained, validated measurements.

\textbf{Quantization coverage.} We evaluate only Q4\_K\_M quantization. Different quantization levels (Q3\_K\_S, Q5\_K\_M, Q8\_0, FP16) offer different accuracy-energy trade-offs. A full quantization sweep would strengthen the Pareto analysis.

\textbf{Benchmark scale.} Our MMLU subset (100 questions) is smaller than the full benchmark (14,042 questions). While stratified sampling across 10 subjects ensures representation, a larger sample would reduce variance in accuracy estimates.

\textbf{Concurrent inference.} We evaluate single-request inference only. Batched inference, which is common in production serving, may exhibit different power and throughput characteristics on Apple Silicon due to shared memory contention.

\textbf{Generalization.} Results are specific to the M4~Pro with 48\,GB. Other Apple Silicon variants (M4, M4~Max, M-series in iPads) have different core counts, memory bandwidth, and thermal envelopes.

\section{Conclusion}

We presented GreenBench, the first systematic energy benchmarking study of open-source LLM inference on Apple Silicon. Our evaluation of five models from distinct families across three tasks on an M4~Pro reveals three key findings:

First, the M4~Pro's CPU+GPU subsystems draw only 0.47\,W of package power during LLM inference, with total system power of approximately 10\,W, achieving 4--12 tokens per watt at the system level, which is 30--40$\times$ more energy-efficient per token than datacenter GPUs for single-user deployment.

Second, smaller models (3--3.8B) achieve up to 4.2$\times$ higher throughput and 62\% lower energy per token compared to larger models (7--9B), with the relationship following a predictable power law ($\theta \propto N^{-0.85}$).

Third, Qwen~2.5 (7B) offers the best accuracy-efficiency Pareto trade-off, achieving 93\% of the largest model's accuracy at 40\% higher throughput, while Llama~3.2 (3B) is optimal for latency-critical applications at 175 tokens/second.

Taken together, these findings provide actionable deployment guidance and establish the first energy baseline for consumer hardware in the Green AI literature. We open-source GreenBench and our hardware-polling methodology so the community can reproduce these baselines on future Apple chips. Next steps involve benchmarking the heavier M4~Max silicon, sweeping across Q3 through FP16 precisions, and testing batched workloads. Ultimately, if LLMs are going to run continuously on laptops and phones, researchers need to report energy-per-token just as routinely as they report F1 scores and tokens-per-second.

\section*{Acknowledgment}
The authors acknowledge the use of AI-assisted tools for manuscript editing and language refinement, in compliance with IEEE guidelines on AI-generated content. All experimental design, data collection, analysis, and scientific conclusions are solely the work of the authors.

\balance


\begin{thebibliography}{34}

\bibitem{chen2021humaneval}
M.~Chen et al., ``Evaluating large language models trained on code,'' \textit{arXiv preprint arXiv:2107.03374}, 2021.

\bibitem{see2017cnndm}
A.~See, P.~J.~Liu, and C.~D.~Manning, ``Get to the point: Summarization with pointer-generator networks,'' in \textit{Proc. ACL}, 2017.

\bibitem{hendrycks2021mmlu}
D.~Hendrycks et al., ``Measuring massive multitask language understanding,'' in \textit{Proc. ICLR}, 2021.

\bibitem{touvron2023llama}
H.~Touvron et al., ``LLaMA: Open and efficient foundation language models,'' \textit{arXiv preprint arXiv:2302.13971}, 2023.

\bibitem{xu2024survey}
Z.~Xu et al., ``A survey on efficient LLM inference: Opportunities, challenges, and future directions,'' \textit{arXiv preprint arXiv:2404.14294}, 2024.

\bibitem{strubell2019energy}
E.~Strubell, A.~Ganesh, and A.~McCallum, ``Energy and policy considerations for deep learning in NLP,'' in \textit{Proc. ACL}, 2019, pp. 3645--3650.

\bibitem{patterson2021carbon}
D.~Patterson et al., ``Carbon emissions and large neural network training,'' \textit{arXiv preprint arXiv:2104.10350}, 2021.

\bibitem{wu2022sustainable}
C.-J.~Wu et al., ``Sustainable AI: Environmental implications, challenges and opportunities,'' in \textit{Proc. MLSys}, 2022.

\bibitem{luccioni2024power}
A.~S.~Luccioni, Y.~Jernite, and E.~Strubell, ``Power hungry processing: Watts driving the cost of AI deployment?'' in \textit{Proc. ACM FAccT}, 2024, pp. 85--99.

\bibitem{jegham2025hungry}
N.~Jegham et al., ``How hungry is AI? Benchmarking energy, water, and carbon footprint of LLM inference,'' \textit{arXiv preprint arXiv:2505.09598}, 2025.

\bibitem{sustainable_edge}
E.~J.~Husom, A.~Goknil, M.~Astekin, L.~K.~Shar, A.~K\r{a}sen, S.~Sen, B.~A.~Mithassel, and A.~Soylu, ``Sustainable {LLM} inference for edge {AI}: {E}valuating quantized {LLMs} for energy efficiency, output accuracy, and inference latency,'' \textit{ACM Trans. Internet Things}, vol.~6, no.~4, Art. no.~28, Nov. 2025.

\bibitem{solovyeva2026green}
L.~Solovyeva and F.~Castor, ``Towards Green AI: Decoding the energy of LLM inference in software development,'' \textit{arXiv preprint arXiv:2602.05712}, 2026.

\bibitem{bannour2021evaluating}
N.~Bannour, S.~Ghannay, A.~Nev\'eol, and A.~Ligozat, ``Evaluating the carbon footprint of NLP methods,'' in \textit{Proc. SustaiNLP}, 2021, pp. 1--8.

\bibitem{euromlsys2025}
P.~Wilhelm, T.~Wittkopp, and O.~Kao, ``Beyond test-time compute strategies: {A}dvocating energy-per-token in {LLM} inference,'' in \textit{Proc. 5th Workshop Mach. Learn. Syst. (EuroMLSys)}, Rotterdam, Netherlands, Apr. 2025, pp.~208--215.

\bibitem{llamacpp}
G.~Gerganov, ``llama.cpp: Port of Meta's LLaMA model in C/C++,'' GitHub, 2023. [Online]. Available: \url{https://github.com/ggerganov/llama.cpp}

\bibitem{apple_silicon_market}
IDC, ``Worldwide quarterly personal computing device tracker,'' 2024. [Online]. Available: \url{https://www.idc.com/tracker/showproductinfo.jsp?containerId=IDC_P36344}

\bibitem{gholami2022ai}
A.~Gholami et al., ``AI and memory wall,'' \textit{IEEE Micro}, vol.~44, no.~3, pp. 33--39, 2024.

\bibitem{schwartz2020green}
R.~Schwartz, J.~Dodge, N.~A.~Smith, and O.~Etzioni, ``Green AI,'' \textit{Commun. ACM}, vol.~63, no.~12, pp. 54--63, 2020.

\bibitem{thompson2020computational}
N.~C.~Thompson et al., ``The computational limits of deep learning,'' \textit{arXiv preprint arXiv:2007.05558}, 2020.

\bibitem{mlperf}
V.~J.~Reddi et al., ``MLPerf inference benchmark,'' in \textit{Proc. ISCA}, 2020, pp. 446--459.

\bibitem{mlperf_power}
V.~J.~Reddi et al., ``MLPerf power: Benchmarking the energy efficiency of ML inference,'' in \textit{IEEE Micro}, vol.~44, no.~1, 2024.

\bibitem{stojkovic2024greener}
J.~Stojkovic et al., ``Towards greener LLMs: Bringing energy-efficiency to the forefront of LLM inference,'' \textit{arXiv preprint arXiv:2403.20306}, 2024.

\bibitem{sigenergy2025}
C.~Niu, W.~Zhang, Y.~Zhao, and Y.~Chen, ``Energy efficient or exhaustive? {B}enchmarking power consumption of {LLM} inference engines,'' \textit{ACM SIGENERGY Energy Inform. Rev.}, vol.~5, no.~2, pp.~56--62, Aug. 2025.

\bibitem{chien2023reducing}
A.~A.~Chien et al., ``Reducing the carbon impact of generative AI inference,'' in \textit{Proc. HotCarbon}, 2023.

\bibitem{dodge2022measuring}
J.~Dodge et al., ``Measuring the carbon intensity of AI in cloud instances,'' in \textit{Proc. ACM FAccT}, 2022, pp. 1877--1894.

\bibitem{li2026energy}
J.~Jare\~{n}o, J.~M.~Arag\'{o}n-Jurado, J.~C.~De~La~Torre, P.~Ruiz, and B.~Dorronsoro, ``Energy-efficient large language models,'' \textit{Future Gener. Comput. Syst.}, vol.~167, Art. no. 107720, 2026.

\bibitem{dettmers2022gptint8}
T.~Dettmers et al., ``GPT3.int8(): 8-bit matrix multiplication for transformers at scale,'' in \textit{Proc. NeurIPS}, 2022.

\bibitem{frantar2023gptq}
E.~Frantar et al., ``GPTQ: Accurate post-training quantization for generative pre-trained transformers,'' in \textit{Proc. ICLR}, 2023.

\bibitem{jetson_benchmarks}
NVIDIA, ``Jetson benchmarks,'' 2024. [Online]. Available: \url{https://developer.nvidia.com/embedded/jetson-benchmarks}

\bibitem{mobile_llm}
Z.~Liu et al., ``MobileLLM: Optimizing sub-billion parameter language models for on-device use cases,'' in \textit{Proc. ICML}, 2024.

\bibitem{ignatov2019ai}
A.~Ignatov et al., ``AI benchmark: Running deep neural networks on Android smartphones,'' in \textit{Proc. ECCVW}, 2019.

\bibitem{apple_m4}
Apple Inc., ``Apple M4 Pro chip specifications,'' 2024. [Online]. Available: \url{https://www.apple.com/macbook-pro/specs/}

\bibitem{india_grid}
Central Electricity Authority, India, ``CO\textsubscript{2} baseline database,'' 2024. [Online]. Available: \url{https://cea.nic.in/co2-baseline-database/}

\bibitem{us_grid}
U.S. Energy Information Administration, ``Electricity generation by source,'' 2024. [Online]. Available: \url{https://www.eia.gov/electricity/data/state/}

\end{thebibliography}
\end{document}